%% file: main.tex
\documentclass[10pt,twocolumn,letterpaper]{article}

\usepackage{iccv}
\usepackage{times}
\usepackage{epsfig}
\usepackage{graphicx}
\usepackage{amsmath}
\usepackage{amssymb}

\usepackage[utf8]{inputenc}
\usepackage{kotex}
\usepackage{kotex-logo}

\usepackage{times}
\usepackage{helvet}
\usepackage{courier}
\usepackage{url}
\usepackage{graphicx}
\usepackage{booktabs}
\usepackage{subcaption}
\usepackage{amsmath}
\usepackage{amsfonts}
\usepackage{mathtools}
\usepackage{multirow}
\usepackage{enumitem}  
\usepackage{pifont}

\usepackage{contour}

\usepackage{algorithm}
\usepackage{algpseudocode}

\usepackage[pagebackref=true,breaklinks=true,colorlinks,bookmarks=false]{hyperref}

\iccvfinalcopy 

\def\iccvPaperID{2580} 

\ificcvfinal\fi

\begin{document}

\title{Conditional Cross Attention Network for Multi-Space Embedding without Entanglement in Only a SINGLE Network}

\author{
Chull Hwan Song$^1$ \ \ \ \ Taebaek Hwang$^1$ \ \ \ \ Jooyoung Yoon$^1$ \ \ \ \ Shunghyun Choi$^1$ \ \ \ \ Yeong Hyeon Gu$^{2}$\thanks{Corresponding author}\\
\normalsize $^1$Dealicious Inc.\quad
\normalsize $^2$Sejong University \\
}

\maketitle
\ificcvfinal\thispagestyle{empty}\fi

\input{abbrev}
\input{defn}

\newcommand{\Note}[3]{{\color{#1}[#2: #3]}}
\definecolor{orange}{rgb}{1,0.5,0}
\newcommand{\song}[1]{\Note{orange}{CH}{#1}}

\definecolor{airforceblue}{rgb}{0.36, 0.54, 0.66}
\newcommand{\schoi}[1]{\Note{airforceblue}{schoi}{#1}}

\definecolor{blue}{rgb}{0,0.5,1}
\newcommand{\baek}[1]{\Note{blue}{TB}{#1}}

%
\input{abstract_en}
\input{intro_en}
\input{relatedworks_en}

\input{methods_en}

\section{Experiments}

\input{datasets_en}
\input{impl_details_en}
\input{vis_en}
\input{exp_memory_en}
\input{sota_en}
\input{ablationstudies_en}
\input{conclusion_en}

{\small
\bibliographystyle{tmlr}
\bibliography{main}
}

\input{supp_en}

\end{document}

%% file: abbrev.tex

\contourlength{0.15pt}

\newcommand{\contourorgange}[1]{%
  \contour{orange}{#1}%
}

\newcommand{\contourgreen}[1]{%
  \contour{green}{#1}%
}

\newcommand\dunderline[3][-1pt]{{%
  \sbox0{#3}%
  \ooalign{\copy0\cr\rule[\dimexpr#1-#2\relax]{\wd0}{#2}}}}

\newcommand{\head}[1]{{\smallskip\noindent\textbf{#1}}}
\newcommand{\alert}[1]{{\color{red}{#1}}}
\newcommand{\sm}{\scriptsize}
\newcommand{\eq}[1]{(\ref{eq:#1})}

\newcommand{\Th}[1]{\textsc{#1}}
\newcommand{\mr}[2]{\multirow{#1}{*}{#2}}
\newcommand{\mc}[2]{\multicolumn{#1}{c}{#2}}
\newcommand{\mca}[3]{\multicolumn{#1}{#2}{#3}}
\newcommand{\tb}[1]{\textbf{#1}}
\newcommand{\ch}{\checkmark}

\newcommand{\red}[1]{{\color{red}{#1}}}
\newcommand{\blue}[1]{{\color{blue}{#1}}}
\newcommand{\green}[1]{\color{green}{#1}}
\newcommand{\gray}[1]{{\color{gray}{#1}}}
\newcommand{\orange}[1]{{\color{orange}{#1}}}

\newcommand{\citeme}[1]{\red{[XX]}}
\newcommand{\refme}[1]{\red{(XX)}}

\newcommand{\fig}[2][1]{\includegraphics[width=#1\linewidth]{fig/#2}}
\newcommand{\figh}[2][1]{\includegraphics[height=#1\linewidth]{fig/#2}}


\newcommand{\tran}{^\top}
\newcommand{\mtran}{^{-\top}}
\newcommand{\zcol}{\mathbf{0}}
\newcommand{\zrow}{\zcol\tran}

\newcommand{\ind}{\mathbbm{1}}
\newcommand{\expect}{\mathbb{E}}
\newcommand{\nat}{\mathbb{N}}
\newcommand{\zahl}{\mathbb{Z}}
\newcommand{\real}{\mathbb{R}}
\newcommand{\proj}{\mathbb{P}}
\newcommand{\prob}{\mathbf{Pr}}
\newcommand{\normal}{\mathcal{N}}

\newcommand{\mif}{\textrm{if}\ }
\newcommand{\other}{\textrm{otherwise}}
\newcommand{\minimize}{\textrm{minimize}\ }
\newcommand{\maximize}{\textrm{maximize}\ }
\newcommand{\st}{\textrm{subject\ to}\ }

\newcommand{\id}{\operatorname{id}}
\newcommand{\const}{\operatorname{const}}
\newcommand{\sgn}{\operatorname{sgn}}
\newcommand{\var}{\operatorname{Var}}
\newcommand{\mean}{\operatorname{mean}}
\newcommand{\trace}{\operatorname{tr}}
\newcommand{\diag}{\operatorname{diag}}
\newcommand{\vect}{\operatorname{vec}}
\newcommand{\cov}{\operatorname{cov}}
\newcommand{\sign}{\operatorname{sign}}
\newcommand{\prj}{\operatorname{proj}}

\newcommand{\sigmoid}{\operatorname{sigmoid}}
\newcommand{\softmax}{\operatorname{softmax}}
\newcommand{\clip}{\operatorname{clip}}

\newcommand{\defn}{\mathrel{:=}}
\newcommand{\peq}{\mathrel{+\!=}}
\newcommand{\meq}{\mathrel{-\!=}}

\newcommand{\floor}[1]{\left\lfloor{#1}\right\rfloor}
\newcommand{\ceil}[1]{\left\lceil{#1}\right\rceil}
\newcommand{\inner}[1]{\left\langle{#1}\right\rangle}
\newcommand{\norm}[1]{\left\|{#1}\right\|}
\newcommand{\abs}[1]{\left|{#1}\right|}
\newcommand{\frob}[1]{\norm{#1}_F}
\newcommand{\card}[1]{\left|{#1}\right|\xspace}
\newcommand{\diff}{\mathrm{d}}
\newcommand{\der}[3][]{\frac{d^{#1}#2}{d#3^{#1}}}
\newcommand{\pder}[3][]{\frac{\partial^{#1}{#2}}{\partial{#3^{#1}}}}
\newcommand{\ipder}[3][]{\partial^{#1}{#2}/\partial{#3^{#1}}}
\newcommand{\dder}[3]{\frac{\partial^2{#1}}{\partial{#2}\partial{#3}}}

\newcommand{\wb}[1]{\overline{#1}}
\newcommand{\wt}[1]{\widetilde{#1}}

\def\xssp{\hspace{1pt}}
\def\ssp{\hspace{3pt}}
\def\msp{\hspace{5pt}}
\def\lsp{\hspace{12pt}}

\hypersetup{
    colorlinks=true,
    linkcolor=black,   
    citecolor=black,   
    urlcolor=black,    
}

\newcommand{\cA}{\mathcal{A}}
\newcommand{\cB}{\mathcal{B}}
\newcommand{\cC}{\mathcal{C}}
\newcommand{\cD}{\mathcal{D}}
\newcommand{\cE}{\mathcal{E}}
\newcommand{\cF}{\mathcal{F}}
\newcommand{\cG}{\mathcal{G}}
\newcommand{\cH}{\mathcal{H}}
\newcommand{\cI}{\mathcal{I}}
\newcommand{\cJ}{\mathcal{J}}
\newcommand{\cK}{\mathcal{K}}
\newcommand{\cL}{\mathcal{L}}
\newcommand{\cM}{\mathcal{M}}
\newcommand{\cN}{\mathcal{N}}
\newcommand{\cO}{\mathcal{O}}
\newcommand{\cP}{\mathcal{P}}
\newcommand{\cQ}{\mathcal{Q}}
\newcommand{\cR}{\mathcal{R}}
\newcommand{\cS}{\mathcal{S}}
\newcommand{\cT}{\mathcal{T}}
\newcommand{\cU}{\mathcal{U}}
\newcommand{\cV}{\mathcal{V}}
\newcommand{\cW}{\mathcal{W}}
\newcommand{\cX}{\mathcal{X}}
\newcommand{\cY}{\mathcal{Y}}
\newcommand{\cZ}{\mathcal{Z}}

\newcommand{\vA}{\mathbf{A}}
\newcommand{\vB}{\mathbf{B}}
\newcommand{\vC}{\mathbf{C}}
\newcommand{\vD}{\mathbf{D}}
\newcommand{\vE}{\mathbf{E}}
\newcommand{\vF}{\mathbf{F}}
\newcommand{\vG}{\mathbf{G}}
\newcommand{\vH}{\mathbf{H}}
\newcommand{\vI}{\mathbf{I}}
\newcommand{\vJ}{\mathbf{J}}
\newcommand{\vK}{\mathbf{K}}
\newcommand{\vL}{\mathbf{L}}
\newcommand{\vM}{\mathbf{M}}
\newcommand{\vN}{\mathbf{N}}
\newcommand{\vO}{\mathbf{O}}
\newcommand{\vP}{\mathbf{P}}
\newcommand{\vQ}{\mathbf{Q}}
\newcommand{\vR}{\mathbf{R}}
\newcommand{\vS}{\mathbf{S}}
\newcommand{\vT}{\mathbf{T}}
\newcommand{\vU}{\mathbf{U}}
\newcommand{\vV}{\mathbf{V}}
\newcommand{\vW}{\mathbf{W}}
\newcommand{\vX}{\mathbf{X}}
\newcommand{\vY}{\mathbf{Y}}
\newcommand{\vZ}{\mathbf{Z}}

\newcommand{\va}{\mathbf{a}}
\newcommand{\vb}{\mathbf{b}}
\newcommand{\vc}{\mathbf{c}}
\newcommand{\vd}{\mathbf{d}}
\newcommand{\ve}{\mathbf{e}}
\newcommand{\vf}{\mathbf{f}}
\newcommand{\vg}{\mathbf{g}}
\newcommand{\vh}{\mathbf{h}}
\newcommand{\vi}{\mathbf{i}}
\newcommand{\vj}{\mathbf{j}}
\newcommand{\vk}{\mathbf{k}}
\newcommand{\vl}{\mathbf{l}}
\newcommand{\vm}{\mathbf{m}}
\newcommand{\vn}{\mathbf{n}}
\newcommand{\vo}{\mathbf{o}}
\newcommand{\vp}{\mathbf{p}}
\newcommand{\vq}{\mathbf{q}}
\newcommand{\vr}{\mathbf{r}}
\newcommand{\Vs}{\mathbf{s}}
\newcommand{\vt}{\mathbf{t}}
\newcommand{\vu}{\mathbf{u}}
\newcommand{\vv}{\mathbf{v}}
\newcommand{\vw}{\mathbf{w}}
\newcommand{\vx}{\mathbf{x}}
\newcommand{\vy}{\mathbf{y}}
\newcommand{\vz}{\mathbf{z}}

\newcommand{\vone}{\mathbf{1}}
\newcommand{\vzero}{\mathbf{0}}

\newcommand{\valpha}{{\boldsymbol{\alpha}}}
\newcommand{\vbeta}{{\boldsymbol{\beta}}}
\newcommand{\vgamma}{{\boldsymbol{\gamma}}}
\newcommand{\vdelta}{{\boldsymbol{\delta}}}
\newcommand{\vepsilon}{{\boldsymbol{\epsilon}}}
\newcommand{\vzeta}{{\boldsymbol{\zeta}}}
\newcommand{\veta}{{\boldsymbol{\eta}}}
\newcommand{\vtheta}{{\boldsymbol{\theta}}}
\newcommand{\viota}{{\boldsymbol{\iota}}}
\newcommand{\vkappa}{{\boldsymbol{\kappa}}}
\newcommand{\vlambda}{{\boldsymbol{\lambda}}}
\newcommand{\vmu}{{\boldsymbol{\mu}}}
\newcommand{\vnu}{{\boldsymbol{\nu}}}
\newcommand{\vxi}{{\boldsymbol{\xi}}}
\newcommand{\vomikron}{{\boldsymbol{\omikron}}}
\newcommand{\vpi}{{\boldsymbol{\pi}}}
\newcommand{\vrho}{{\boldsymbol{\rho}}}
\newcommand{\vsigma}{{\boldsymbol{\sigma}}}
\newcommand{\vtau}{{\boldsymbol{\tau}}}
\newcommand{\vupsilon}{{\boldsymbol{\upsilon}}}
\newcommand{\vphi}{{\boldsymbol{\phi}}}
\newcommand{\vchi}{{\boldsymbol{\chi}}}
\newcommand{\vpsi}{{\boldsymbol{\psi}}}
\newcommand{\vomega}{{\boldsymbol{\omega}}}

\newcommand{\rLambda}{\mathrm{\Lambda}}
\newcommand{\rSigma}{\mathrm{\Sigma}}

\newcommand{\vLambda}{\bm{\rLambda}}
\newcommand{\vSigma}{\bm{\rSigma}}

\makeatletter
\newcommand*\bdot{\mathpalette\bdot@{.7}}
\newcommand*\bdot@[2]{\mathbin{\vcenter{\hbox{\scalebox{#2}{$\m@th#1\bullet$}}}}}
\makeatother

\makeatletter
\DeclareRobustCommand\onedot{\futurelet\@let@token\@onedot}
\def\@onedot{\ifx\@let@token.\else.\null\fi\xspace}

\def\eg{\emph{e.g}\onedot} \def\Eg{\emph{E.g}\onedot}
\def\ie{\emph{i.e}\onedot} \def\Ie{\emph{I.e}\onedot}
\def\cf{\emph{cf}\onedot} \def\Cf{\emph{Cf}\onedot}
\def\etc{\emph{etc}\onedot} \def\vs{\emph{vs}\onedot}
\def\wrt{w.r.t\onedot} \def\dof{d.o.f\onedot} \def\aka{a.k.a\onedot}
\def\etal{\emph{et al}\onedot}
\makeatother

%% file: defn.tex

\newcommand{\ours}{CCA\xspace}
\newcommand{\Ours}{\emph{deep token pooling} (\ours)\xspace}


\newcommand{\cls}{{\texttt{[CLS]}}\xspace}
\newcommand{\patch}{{\texttt{[PATCH]}}\xspace}
\newcommand{\pos}{{\texttt{[POS]}}\xspace}

\newcommand{\relu}{\operatorname{relu}}
\newcommand{\conv}{\operatorname{conv}}
\newcommand{\aconv}{\operatorname{aconv}}

\newcommand{\fc}{\textsc{fc}}
\newcommand{\gap}{\textsc{gap}}
\newcommand{\bn}{\textsc{bn}}
\newcommand{\dropout}{\textsc{dropout}}

\newcommand{\elm}{\textsc{elm}}
\newcommand{\irb}{\textsc{irb}}
\newcommand{\wav}{\textsc{wb}}
\newcommand{\aspp}{\textsc{aspp}}
\newcommand{\fuse}{\textsc{fuse}}


\def\oxf5k{Ox5k\xspace}
\def\paris6k{Par6k\xspace}
\def\roxf{$\cR$Oxford\xspace}
\def\rox{$\cR$Oxf\xspace}
\def\ro{$\cR$O\xspace}
\def\rpar{$\cR$Paris\xspace}
\def\rpa{$\cR$Par\xspace}
\def\rp{$\cR$P\xspace}
\def\r1m{$\cR$1M\xspace}
\def\rs{$\cR$100k\xspace}


\newcommand{\gain}[1]{{\color{green!60!black}#1}}

%% file: abstract_en.tex
\begin{abstract}

Many studies in vision tasks have aimed to create effective embedding spaces for single-label object prediction within an image. However, in reality, most objects possess multiple specific attributes, such as shape, color, and length, with each attribute composed of various classes. To apply models in real-world scenarios, it is essential to be able to distinguish between the granular components of an object. Conventional approaches to embedding multiple specific attributes into a single network often result in entanglement, where fine-grained features of each attribute cannot be identified separately.
To address this problem, we propose a Conditional Cross-Attention Network that induces disentangled multi-space embeddings for various specific attributes with only a single backbone. Firstly, we employ a cross-attention mechanism to fuse and switch the information of conditions (specific attributes), and we demonstrate its effectiveness through a diverse visualization example. Secondly, we leverage the vision transformer for the first time to a fine-grained image retrieval task and present a simple yet effective framework compared to existing methods. Unlike previous studies where performance varied depending on the benchmark dataset, our proposed method achieved consistent state-of-the-art performance on the FashionAI, DARN, DeepFashion, and Zappos50K benchmark datasets.

\end{abstract}



%% file: intro_en.tex
\begin{figure}[t]
\centering\includegraphics[width=0.85\columnwidth]{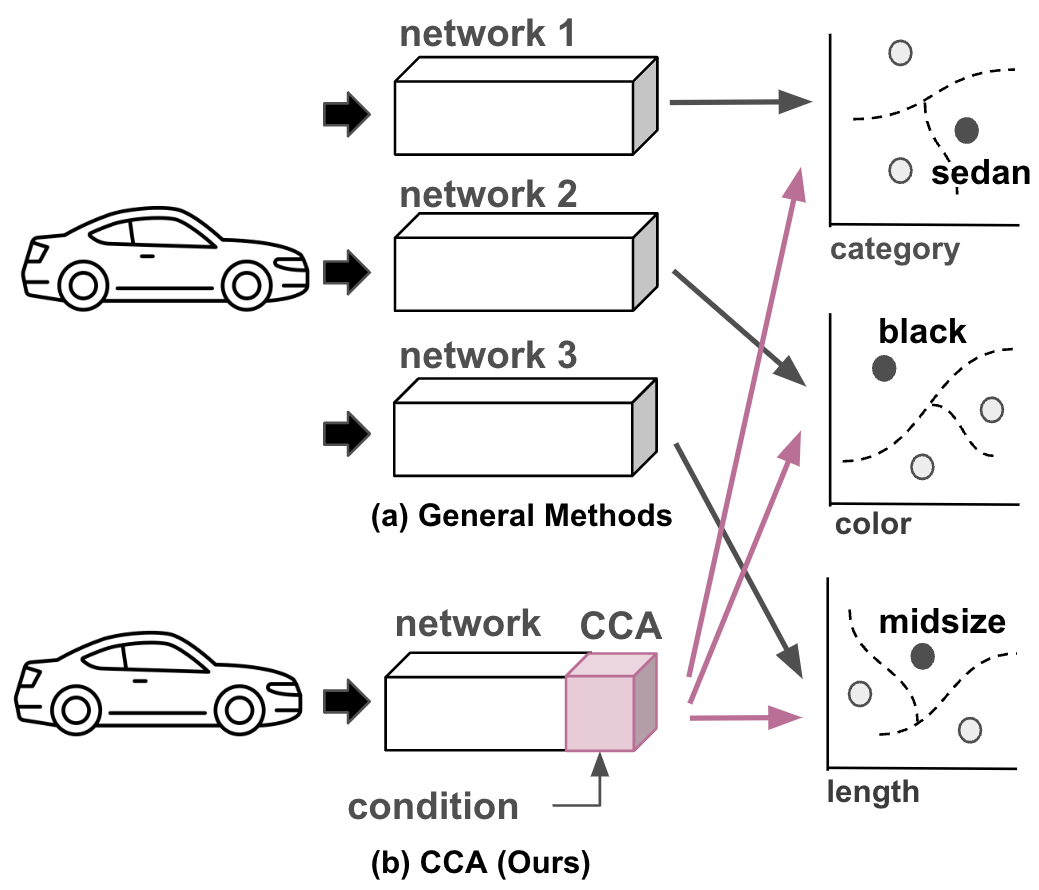}
\vspace{-7pt}
   \caption{ Multiple Networks vs Single Network for Multi-space embedding. CCA means our proposed Conditional Cross Attention Network. 
   }
\label{fig:fig1}
\end{figure}

\section{Introduction}

\begin{figure*}
\begin{center}
\includegraphics[width=0.95\linewidth]{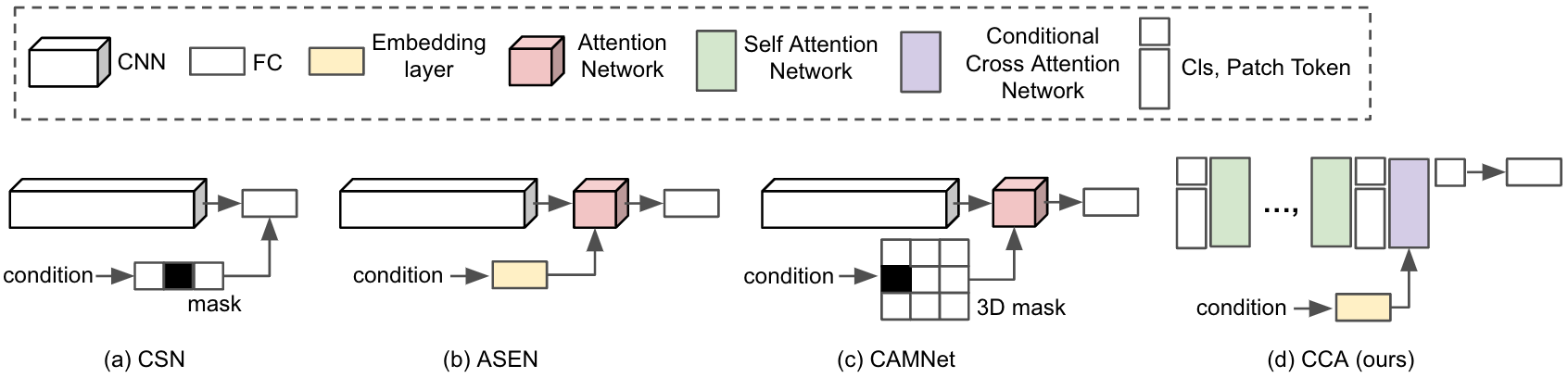}
\end{center}
   \vspace{-16pt}
   \caption{ Previous works (CSN, ASEN, CAMNet) \vs Ours (CCA) }
\label{fig:fig2}
\end{figure*}

ImageNet~\cite{imagenet21k} is a representative benchmark dataset to verify the visual feature learning effects of deep learning models in the vision domain. However, each image has only one label, which cannot fully explain the various features of real objects. For example, a car can be identified with various attributes such as category, color, and length, as in \autoref{fig:fig1}. As shown in \autoref{fig:fig1} (a), the general method of forming embeddings for objects’ various attributes involves constructing neural networks equal to the number of specific attributes, and creating multiple embeddings for vision tasks such as image classification \cite{ResNet_He_2016_CVPR, efficientnet_pmlr_tan_19, Hu01} and retrieval \cite{Kalantidis01, SCH01}. Unlike conventional methods, this study presents a technique that embeds various attributes into a single network. We refer to this technique as multi-space attribute-specific embedding \autoref{fig:fig1} (b).

Embedding space aims to encapsulate feature similarities by mapping similar features to close points and dissimilar ones to farther points. However, when the model attempts to learn multiple visual and semantic concepts simultaneously, the embedding space becomes complex, resulting in entanglement; thus, points corresponding to the same semantic concept can be mapped in different regions. Consequently, embedding multiple concepts in an image into a single network is very challenging. Although previous studies attempted to solve this problem using convolutional neural networks (CNNs) \cite{veit2017, ma2020fine, dong2021fine, SCH01}, they have required intricate frameworks, such as the incorporation of multiple attention modules or stages, in order to identify specific local regions that contain attribute information.

Recently, there has been an increase in research related to ViT \cite{VIT}, which outperforms existing CNN-based models in various vision tasks, such as image classification \cite{VIT}, retrieval \cite{dtop}, and detection \cite{detr}. In addition, research analyzing how ViT learns representations compared to CNN is underway \cite{raghu2021vision, park2022vision, naseer2021intriguing}. Raghu \etal \cite{raghu2021vision} demonstrated that the higher layers of ViT are superior in preserving spatial locality information, providing improved spatially discriminative representation than CNN. Some attributes of an object are more easily distinguished when focusing on specific local areas. So, we tailor the last layer of ViT to recognize specific attributes based on their spatial locality, which provides fine-grained information about a particular condition. \autoref{fig:fig2} summarizes the difference between existing CNN-based and proposed ViT-based methods. This study makes the following contributions:
\begin{enumerate}[itemsep=2pt, parsep=0pt, topsep=0pt]
    \item Entanglement occurs when embedding an object containing multiple attributes using a single network. The proposed CCA that applies a cross-attention mechanism can solve this problem by adequately fusing and switching between the different condition information (specific attributes) and images. 
        
    \item This is the first study to apply ViT to multi-space embedding-based image retrieval tasks. In addition, it is a simple and effective method that can be applied to the ViT architecture with only minor modification. Moreover, it improves memory efficiency by forming multi-space embeddings with only one ViT backbone rather than multiple backbones.
    
    \item Most prior studies showed good performance only on specific datasets. However, the proposed method yields consistently high performance on most datasets and effectively learns interpretable representations. Moreover, the proposed method achieved state-of-the-art (SOTA) performance on all relevant benchmark datasets compared to existing methods.
\end{enumerate}


%% file: relatedworks_en.tex
\section{Related Works}

\paragraph{Similarity Embedding} Triplet Network \cite{van01, Schroff_2015_CVPR} uses distance calculation to embed images into a space; images in the same category are placed close and those in different categories are far apart. This algorithm has been widely used for diverse subjects such as face recognition and image retrieval. However, as it learns from a single embedding space, it is unsuitable for embedding multiple subjects with multiple categories. Multiple learning models must be created separately according to the number of categories to increase the sophistication level. 
\vspace*{-10pt}
%
\paragraph{Image Retrieval via CNN-based Embedding} Image Retrieval is a common task in computer vision, which is finding relevant images based on a query image. Recent works have explored the CNN-based embedding and attention mechanisms to improve image retrieval. Some works leverage attention mechanisms according to the channel-wise \cite{Hu01, wang01, Woo01} and spatial-wise \cite{Woo01} concepts to assign more importance to attended object in the image. Understanding the detailed characteristics of objects is crucial in image retrieval. This is particularly significant in the fashion domain, where even the same type of clothing can have various attributes such as color, material, and length. Therefore, to excel in attribute-based retrieval,, it is required to recognize disentangled representation for each attribute. The nature of this task is suitable for demonstrating the effectiveness of multi-space embedding. Thus, we show the efficacy of CCA through a fashion attribute-specific retrieval task.

\vspace*{-10pt}
\paragraph{CNN based Attributes-Specific Embedding} 
\autoref{fig:fig2} outlines the concepts of existing attribute-specific embedding, similar to our current study. CSN \cite{veit2017} converts the condition into a mask-like representation for multi-space embedding. The mask can be easily applied to the fully connected layer (FC).
ASEN \cite{ma2020fine} joins the attention mechanism with a condition for multi-space embedding. A variation, ASEN++ \cite{dong2021fine}, extended ASEN to 2 stages. These multi-stage techniques are excluded from this study for a fair comparison. 
M2Fashion \cite{Wan01} adds a classifier to the ASEN base.
Unlike CSN, CAMNet \cite{song2022} was extended to 3D feature maps and applied to the spatial attention mechanism, thus enhancing performance. These studies are CNN-based, not self-attention-based like the present study. The recent ViT \cite{VIT} has been successfully applied to many vision tasks. However, there has been no technique of multi-space embedding for specific attributes, as described in this study.

%% file: methods_en.tex
\section{Methods}

\begin{figure*}
\begin{center}
\includegraphics[width=0.85\linewidth]{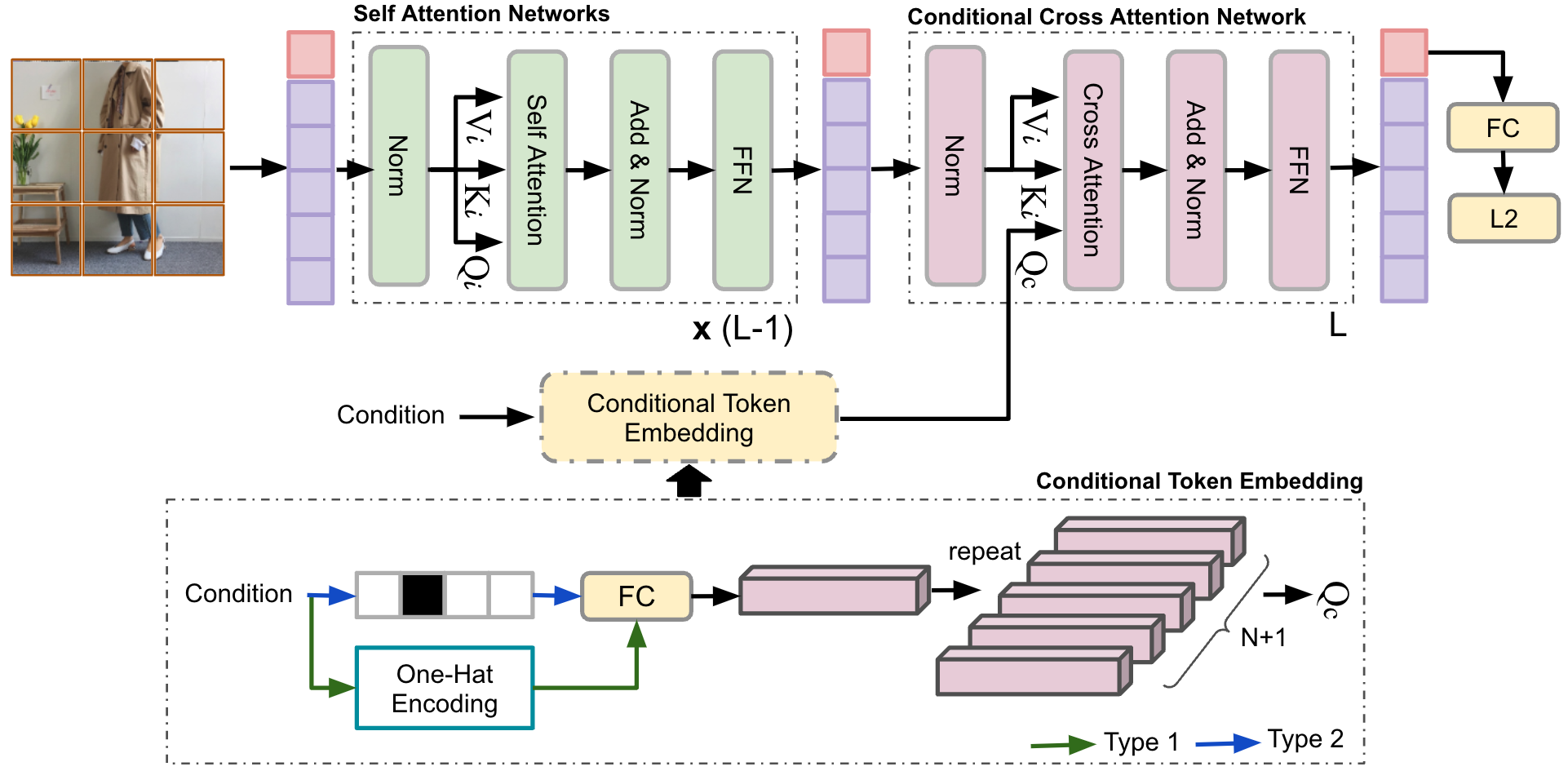}
\end{center}
   \vspace{-18pt}
   \caption{ The Architecture of Conditional Cross Attention Network (CCA) }
\label{fig:fig3}
\end{figure*}

\autoref{fig:fig3} presents the proposed CCA architecture, which is mostly similar to that of ViT \cite{VIT} because it was designed to embed specific attributes through a detailed analysis of the ViT architecture. Hence, CCA is easily applicable under the ViT architecture. Moreover, as described in \autoref{sec:eval}, it yields excellent performance. The proposed architectures comprise self-attention and CCA modules. The following sections explain these networks.
\subsection{Self Attention Networks}
\label{sec:ssn}
The self-attention module learns a common representation containing the information necessary for multi-space embedding. The self-attention modules are nearly identical to ViT \cite{VIT}. ViT divides the image into specific patch sizes and converts it into continuous patch tokens (\patch). Here, classification tokens (\cls) \cite{bert} are added to the input sequence. As self-attention in ViT is position-independent, position embeddings are added to each patch token for vision applications requiring position information. All tokens of ViT are forwarded through a stacked transformer encoder and used for classification using \cls of the last layer. The transformer encoder consists of feed-forward (FFN) and multi-headed self-attention (MSA) blocks in a continuous chain. FFN has two multi-layer perceptrons; layer normalization (LN) is applied at the beginning of the block, followed by residual shortcuts. The following equation is for the $l$-th transformer encoder. 

\begin{equation} 
\begin{split}
\label{eq:vit}
    \mathbf{x}_{0} & = \left[ \mathbf{x}_{\cls} ; \mathbf{x}_{\patch} \right] + \mathbf{x}_{\pos} \\
    \mathbf{x^\prime}_l & = \mathbf{x}_{l-1} + \mathtt{MSA}(\mathtt{LN}(\mathbf{x}_{l-1})) \\
    \mathbf{x}_l & = \mathbf{x^\prime}_l + \mathtt{FFN}(\mathtt{LN}(\mathbf{x^\prime}_l))
 \end{split}
\end{equation}

where $\mathbf{x}_{0}$ is inital ViT input. $\mathbf{x}_{\cls} \in \mathbb{R}^{1\times D}$, $\mathbf{x}_{\patch} \in \mathbb{R}^{N\times D}$ and $\mathbf{x}_{\pos} \in \mathbb{R}^{(1+N)\times D}$ are the classification, patch, and positional embedding, respectively. The output of the $L-1$ repeated encoder is used as input to the CCA module, as explained in \autoref{sec:cca}

\begin{figure*}
\begin{center}
\includegraphics[width=0.95 \linewidth]{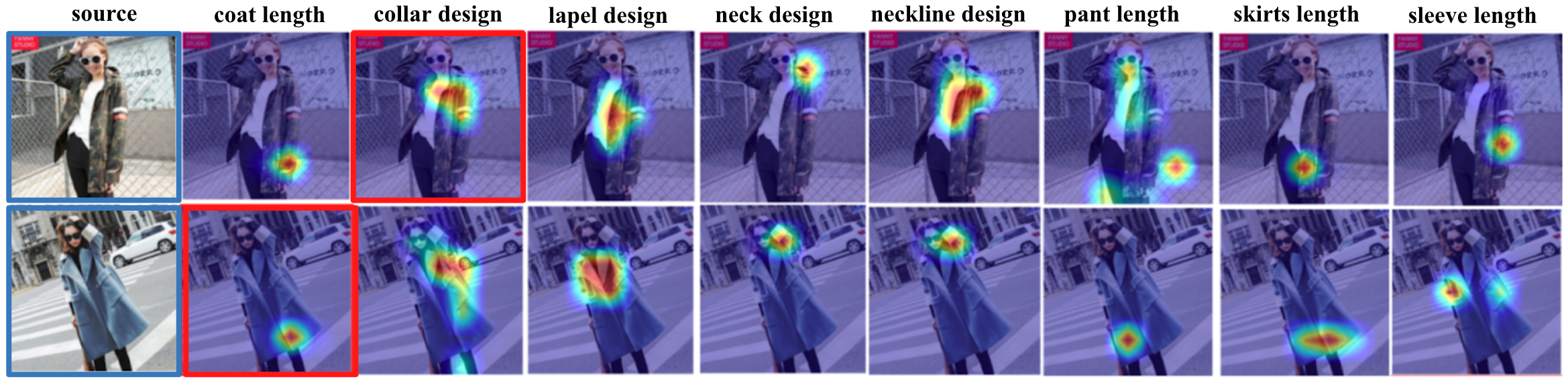}
\end{center}
\vspace{-18pt}
   \caption{Visualization of attention heat maps for each attribute. Red outlines denote actual annotated attributes in FashionAI.}
\label{fig:fig6}
\end{figure*}

\subsection{Conditional Cross Attention Network}
\label{sec:cca}
In this study, the transformer must fuse the concept of attributes and mapped condition information for the network to learn. Drawing inspiration from Vaswani \etal \cite{Transformer_NIPS2017_Vaswani}, we propose CCA to enable learning in line with the transformer's self-attention mechanism. CCA uses a common representation obtained from the self-attention module and cross-attention of the mask according to the given condition to learn nonlinear embeddings that effectively express the semantic similarity based on the condition. Though existing techniques, such as CSN \cite{veit2017} and ASEN \cite{ma2020fine}, have applied condition information to the embedding, these methods are CNN-based rather than transformer-based. 
\vspace*{-10pt}
\paragraph{Conditional Token Embedding} 
Some network switch based on the condition is needed to embed multiple attributes under a single network. 
In other words, attributes must be learned according to the condition. This study proposes two conditional token embedding methods, as shown in  \autoref{fig:fig3}.

First, Condition ${c}$ is converted into a one-hot vector form, after which conditional token embedding is performed, similar to that used in multi-modal studies such as DeViSE \cite{DeViSE}, which learns text and image information having the same meaning using heterogeneous data in the same space, as follows:

\begin{equation}
\label{eq:Conditionalembedding1}
   {\mathbf{q}_\mathbf{c}} = \mathtt{FC(onehot({c}))} 
\end{equation}

where $\mathbf{q}_\mathbf{c} \in \mathbb{R}^{D \times 1 }$, ${c}$ is condition of size $K$.

Second is the CSN \cite{veit2017} technique, presented in \autoref{fig:fig2} (a). To express $K$ conditions, CSN applies a mask $\in \mathbb{R}^{K\times D}$ to one of the features and uses element-wise multiplication to fuse and embed two CNN features $\in \mathbb{R}^{D}$. This study uses this step only for conditional feature embedding without fusing the features. To this end, we initialize the mask $ \in \mathbb{R}^{K \times D}$ for all attributes. This mask can be expressed as a learnable lookup table. The conditional token embedding using the mask is expressed as follows:
\begin{equation}
\begin{array}{l}
{\mathbf{q}_\mathbf{c}} = \mathtt{FC}(\phi(\mathbf{M_{\theta}}{[c, :]}))
\label{eq:Conditionalembedding2}
\end{array} 
\end{equation}

where $\phi$ refers to ReLU, the activation function. Accordingly, the dimensions must be the same as the feature to apply self-attention. The result of $\mathbf{FC}$ in \autoref{eq:Conditionalembedding1} and \autoref{eq:Conditionalembedding2} is embedded while matching the dimension of $C$.

Finally, the result of both equations must equal the dimensions of the token embedding in \autoref{sec:ssn}. Therefore, the same vector $\mathbf{q}_\mathbf{c} \in \mathbb{R}^{D \times 1 }$ is repeated times to expand the result of both equations as follows:

\begin{equation}
\begin{array}{l}
{Q_c} = [\mathbf{q}_\mathbf{c} ; \mathbf{q}_\mathbf{c} ; ... ; \mathbf{q}_\mathbf{c}]
\label{eq:Conditionalembedding3}
\end{array} 
\end{equation}

\vspace*{-10pt}

\paragraph{Conditional Cross Attention}
Finally, the transformer architecture must effectively fuse the conditional token embedding vector $Q_c$, for which we use CCA.
The MSA process in \autoref{eq:vit} uses a self-attention mechanism with the vector query ($Q$), key ($K$), and Value ($V$) as input and is expressed as follows:
 
\begin{equation}
\label{eq:selfattention}
   \mathtt{Attention}{(Q_i, K_i, V_i)} = \mathtt{softmax}{(\frac{Q_{i}K_{i}^\top}{\sqrt{d}})V_i}
\end{equation}

These vectors generated from image $i$ can be expressed using ${K_{i}, Q_i, V_i} \in \mathbb{R}^{N \times D}$, consistent with the tokens mentioned above. The inner product of ${Q}$ and ${K}$ is calculated, which scales and normalizes with the softmax function to obtain weight N.

In contrast, though CCA is nearly identical to self-attention, Query, $Q_c$ in \autoref{eq:Conditionalembedding3}, is generated to have condition information. $K_i$ and $V_i$, which are the same as above, are input, and the cross-attention mechanism is applied to construct the final CCA as follows. 

\begin{equation}
\label{eq:crossattention}
   \mathtt{Attention}{(Q_c, K_i, V_i)} = \mathtt{softmax}{(\frac{Q_{c}K_{i}^\top}{\sqrt{d}})V_i}
\end{equation}

The cross-attention mechanism is nearly identical to general self-attention; except for the part of \autoref{eq:crossattention}, it is the same as \autoref{eq:vit}. 
The output is the embedding values of \cls and \patch. In our proposed CCA, only \cls is used for the loss calculation. For the final output, FC and l2 normalization are applied to the embedding feature $x_{\cls} \in \mathbb{R}^{D}$ of \cls as follows:

\begin{equation}
\label{eq:finalcls}
    f^{final}= \mathtt{l{2}}(\mathtt{FC}{(x_{\cls})})
\end{equation}

Self-attention, explained in \autoref{sec:ssn}, executes the transformer encoder until step $1 \sim (L-1) $, while CCAN, explained in \autoref{sec:cca}, applies only to the final step $L$. In other words, during inference, as shown in \autoref{fig:fig1}, if step $1 \sim (L-1)$ is executed only once and the condition in the final step $L$ is changed and repeated, then several specific features can be obtained under various conditions. \autoref{fig:fig6} shows related experimental results. Eight attributes in the FashionAI dataset are attended in regions matched to each attribute. In addition, step $1 \sim (L-1) $ in the network model can apply the existing ViT-based pre-trained model without modification for learning.

\subsection{Triplet Loss with Conditions}
\label{sec:triplet}
We use triplet loss for learning specific attributes, different from the previous general triplet loss in that a conditioned triplet must be constructed. If a label with image $I$ and condition $c$ exists, then the Pair can be denoted as $(I, L_{c})$. When expanded to triplets, this is expressed as follows.

\begin{equation}
\label{eq:loss1}
    \mathcal{T}= \{((I^a,L^{a}_c), (I^+,L^{+}_c), (I^-,L^{-}_c)| c)\}
\end{equation}

where $a$ indicates the anchor, $+$ means that it has the same class in the same condition as the anchor, and $-$ means that it does not have the same class. Using negative samples with the same condition in triplet learning can be interpreted as a hard negative mining strategy. As shown in \cite{xuan2020hard}, randomly selected negatives are easily distinguished from anchors, enabling the model to learn only coarse features. However, for negative samples with the same condition, the model must distinguish more fine-grained differences. Hence, informative negative samples more suitable for specific-attributes learning are provided. The equation of triplet loss $\mathcal{L}$ is as follows.

\begin{equation}
\label{eq:loss2}
\begin{multlined}
\mathcal{L}(I^a, I^+, I^- , c) = \\ \max\{0, \textsc{Dist}(I^a,I^+|c) - \textsc{Dist}(I^a,I^-|c) + m\}
\end{multlined}
\end{equation}

$m$ uses a predefined margin, and $\textsc{Dist()}$ refers to cosine distance. In the Appendix, we present Algorithm \autoref{alg_train}, which outlines the pseudo-code of our proposed method.

%% file: datasets_en.tex
\autoref{tab:datasets} shows the statistics of the datasets, including the number of attributes, classes within the attributes, and the total number of images. The difficulty increases as the number of attributes increases and for higher classes. These results can also be seen in the evaluation results of \autoref{tab:fashionAI}, \autoref{tab:DARN}, and \autoref{tab:DeepFashion}. 
%

\begin{table} [tb!]
\renewcommand{\arraystretch}{1.0}
\centering 
\resizebox{0.9\columnwidth}{!}{
\begin{tabular}{lccc}
\toprule
\textbf{DataSets} & \textbf{\#Attributes} & \textbf{\#Classes} & \textbf{\#Images} \\
\midrule
FashionAI \cite{Zou2019FashionAIAH} & 8 & 55 & 180,335\\
DARN \cite{Huang01} & 9 & 185 & 195,771 \\
DeepFashion \cite{liu2016deepfashion} & 6 & 1000 & 289,222\\
Zappos50k \cite{Zappos} & 4 & 34 & 50,025\\
\bottomrule
\end{tabular}
 }
 \vspace{-8pt}
 \caption{Statistics of the banchmark datasets.}
\label{tab:datasets}
\end{table}

\subsection{Metrics}
\label{sec:Metrics} 
For FashionAI, DARN, and DeepFashion, we used the experimental setting information of ASEN \cite{ma2020fine} and applied the mean average precision (mAP) metric for evaluation. For Zappos50K, we followed the experimental setting of CSN \cite{veit2017} and applied the triplet prediction metric for evaluation. This metric verifies the efficiency of attribute specific embedding learning for predicting triplet relationships.

%% file: impl_details_en.tex
\subsection{Implementation Details}
\label{sec:impldetails}

The experimental environment was implemented using 8 RTX 3090 GPUs. We used Pytorch \cite{Paszke01} for all implementations. 
The backbone network was initialized with pre-trained R50+ViT-B/16 \cite{VIT}.
A batch size of 64 and learning rate of 0.0001 was applied for learning. We trained the models up to 200 epochs and selected the trained model that yielded the best results. Triplet loss, described in \autoref{sec:triplet}, was used with a margin of 0.2.

%% file: vis_en.tex
\begin{figure*}
\begin{center}
\includegraphics[width=1\linewidth]{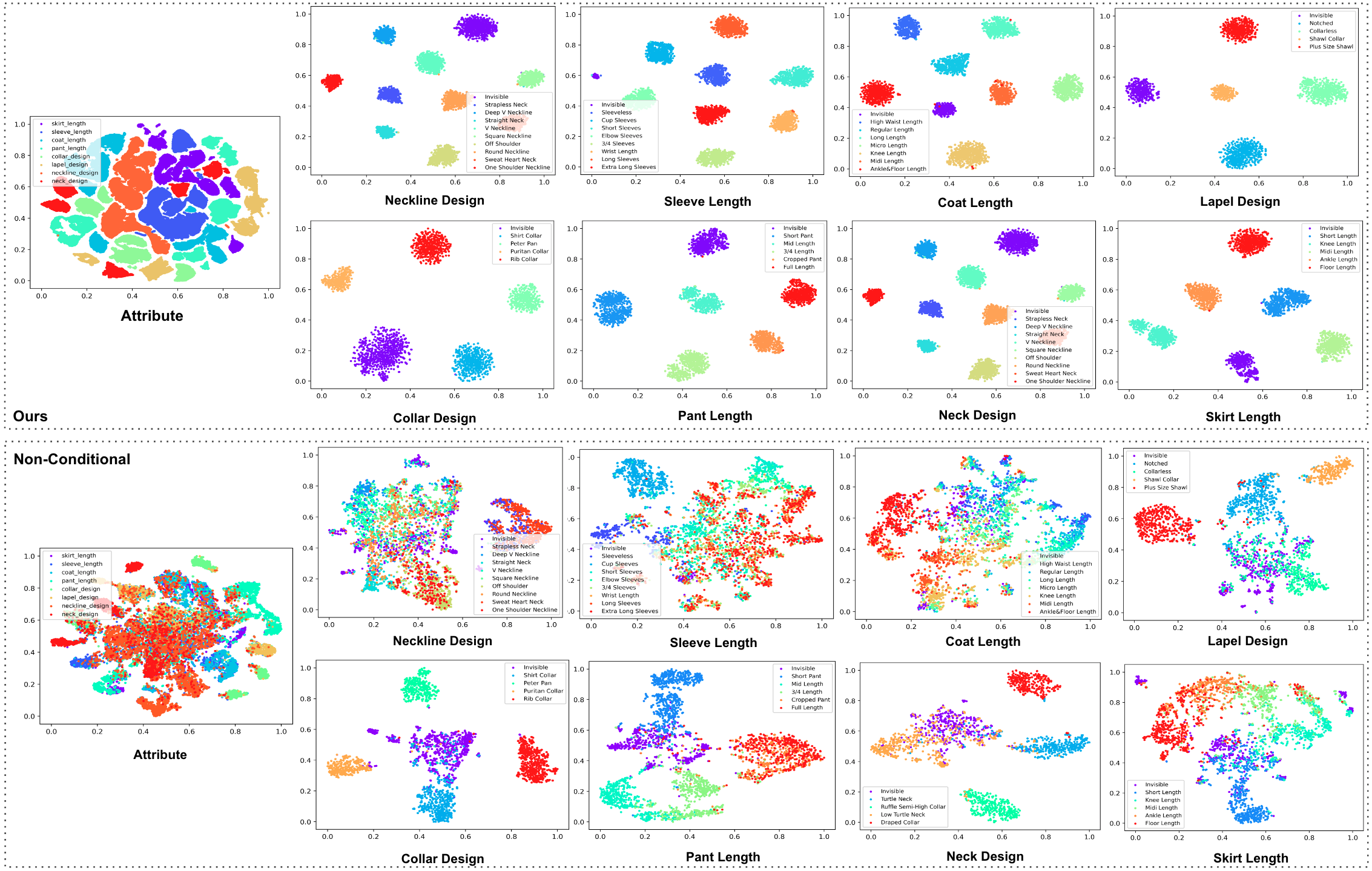}
\end{center}
\vspace{-17pt}
   \caption{Comparison of multi-space embedding : Conditional (Ours) \vs Non-Conditional (Triplet network)}
\label{fig:fig4}
\end{figure*}

\input{sota_fashionai.tex}

\subsection{Visualization of Multi-Space Embedding and Ranking Results}
\label{sec:vis}

\paragraph{Entangled \vs Disentangled Multi-space Embedding} 
The proposed method enables multi-space embedding for various specific attributes with only one backbone network. When using the general learning method, entanglement in the embedding space inevitably occurs. To solve the entanglement problem and verify whether multi-space embeddings were formed, t-SNE \cite{Laurens01} was used to examine the results. The t-SNE visualization results in \autoref{fig:fig4} show whether each attribute class of the FashionAI dataset is properly embedded. The t-SNE visualization results at the center are for the FashionAI dataset with 8 fashion attributes. For the proposed method, excellent embedding results are found for all 8 attributes in the center, and each attribute on the edges. However, training a single model for multiple attributes with the non-conditional method, which is the triplet network in \autoref{tab:fashionAI}, \autoref{tab:DARN}, and \autoref{tab:DeepFashion}, do not solve the entanglement problem. These findings offer strong evidence that the proposed method achieved multi-space embedding with only one backbone network.

\begin{figure*}
\begin{center}
\includegraphics[width=0.8\linewidth]{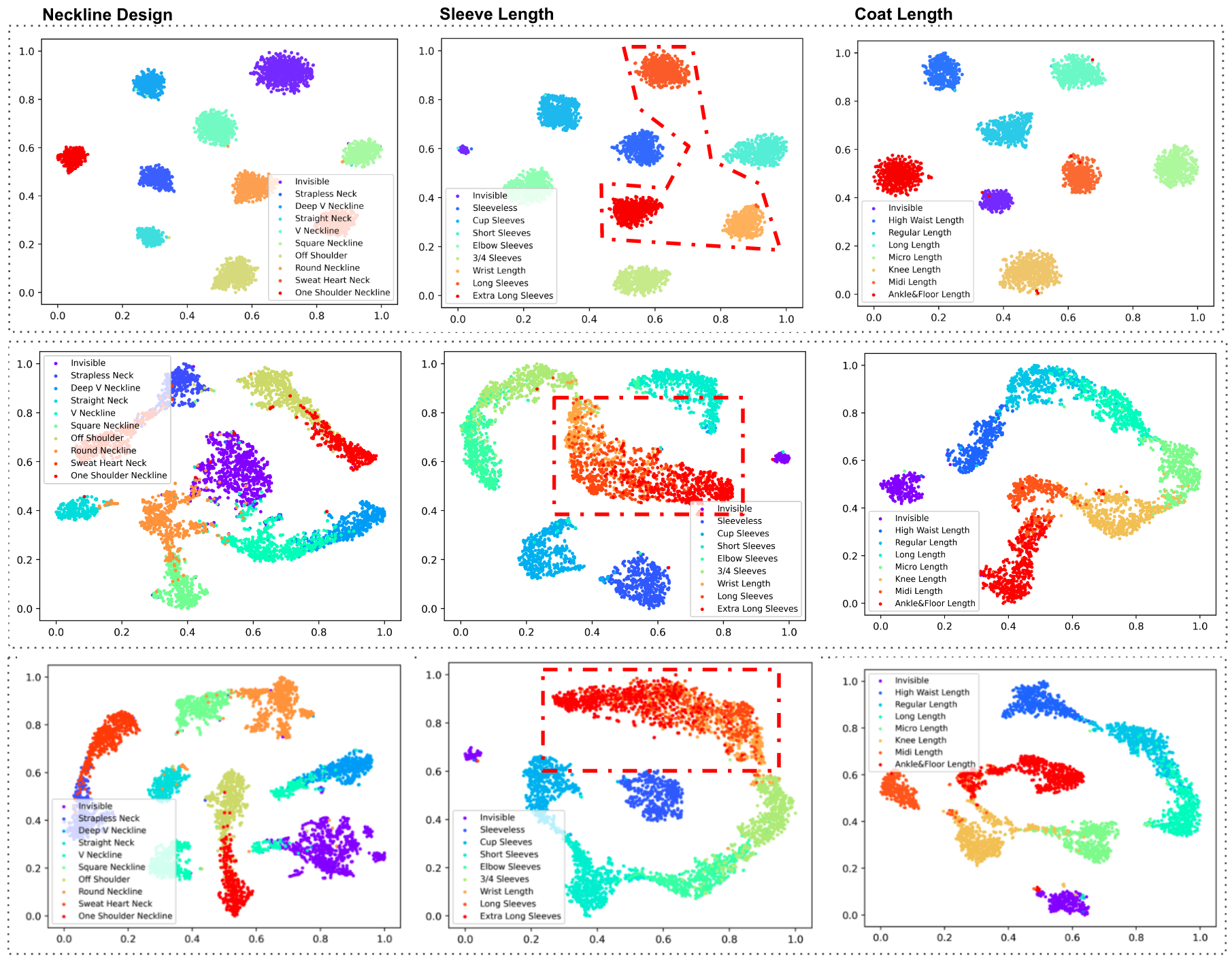}
\end{center}
\vspace{-18pt}
    \caption{Comparison of multi-space embeddings (Ours vs. ASEN and CAMNet) for FashionAI. The top row corresponds to our method, the middle row to ASEN, and the bottom row to CAMNet. The embeddings are shown for three categories (Neck Design, Sleeve Length, Coat Length) out of eight attributes.
   }
\label{fig:fig7}
\end{figure*}
\vspace*{-10pt}
\paragraph{Ours \vs Previous Works' Multi-space Embedding} 

\autoref{fig:fig7} compares the embedding results between the proposed and previous (ASEN~\cite{ma2020fine}, CAMNet~\cite{song2022}) methods for the FashionAI dataset. The comparison results for 3 of the 8 detailed categories (Neck Design, Sleeve Length, and Coat Length) in FashionAI are shown. Our method yielded better embedding results than ASEN and CAMNet. For example, in the ASEN and CAMNet results, entanglement occurred in the embedding space for the Wrist Length, Long Sleeves, and Extra-long Sleeves classes of Sleeve Length, whereas entanglement is resolved with our proposed method. \autoref{fig:app4} presents the embedding results for the 8 attributes in the FashionAI dataset. 

\vspace*{-10pt}
\paragraph{Ranking Results} 
\autoref{fig:app3} in \autoref{sec:more_vis} presents the Top 3 ranking results for the 8 attributes in the FashionAI dataset. The order in the figure is lapel design (notched), neckline design (round), skirt length (floor), pant length (midi), sleeve length (short), neck design (low turtle), coat length (midi), and collar design (peter pan). The features of each attribute are reflected accurately in the ranking. This is also demonstrated in the attention heat map.

%% file: sota_fashionai.tex
\begin{table*} [tb!]
\renewcommand{\arraystretch}{1.25} 
\centering 
\resizebox{2\columnwidth}{!}
{
\begin{tabular}{l*{10}{c}}
\toprule
\multirow{2}{*}{{\tb{Method}}} & \multirow{2}{*}{{\tb{Backbone}}} & \multirow{2}{*}{{\tb{mAP}}}& \multicolumn{8}{c}{{\tb{mAP for each attribute}}}  \\
\cmidrule(l){4-11}
 & & &  \tb{skirt length} & \tb{sleeve length} & \tb{coat length} & \tb{pant length} & \tb{collar design} & \tb{lapel design} & \tb{neckline design} & \tb{neck design} \\
\toprule
Random baseline \cite{ma2020fine}  &  R50  &  15.79 & 17.20 & 12.50 & 13.35 & 17.45 & 22.36 & 21.63 & 11.09 & 21.19 \\
Triplet network \cite{ma2020fine}  &  R50  & 38.52 & 48.38 & 28.14 & 29.82 & 54.56 & 62.58 & 38.31 & 26.64 & 40.02  \\
CSN  \cite{ma2020fine}    &  R50 & 53.52 & 61.97 & 45.06 & 47.30 & 62.85 & 69.83 & 54.14 & 46.56 & 54.47  \\
ASEN \cite{ma2020fine}  & R50 & {61.02} & {64.44} & {54.63} & {51.27} & {63.53} & {70.79} & {65.36} & {59.50} & {58.67} \\
CAMNet \cite{song2022}  & R50   & \color{black}{{61.97}} & 64.14 & \color{black}{{56.22}} & \color{black}{{53.05}} &  \color{black}{{65.67}} & \color{black}{{72.60}} & \color{black}{67.74} & \color{black}{{63.05}} &  \color{black}{{61.97}} \\
ASEN++ \cite{dong2021fine} & R50 &  64.31 & 66.34 & 57.53 & 55.51 & 68.77 & 72.94 & 66.95 & 66.81 & \textbf{67.01} \\

TF-CSN$^\dagger$  & ViT & \color{black}{{64.86}} & \color{black}{{66.73}} & \color{black}{{59.58}} & \color{black}{{59.94}} & \color{black}{{70.91}} & \color{black}{{71.45}} & \color{black}{{68.17}} & \color{black}{{64.92}} & \color{black}{{62.33}} \\
TF-ASEN$^\dagger$  & ViT & \color{black}{{64.21}} & \color{black}{{65.86}} & \color{black}{{60.11}} & \color{black}{{59.74}} & \color{black}{{70.20}} & \color{black}{{70.80}} & \color{black}{{67.01}} & \color{black}{{64.08}} & \color{black}{{59.48}} \\
\toprule
\multicolumn{10}{c}{{\tb{Ours }}} \\
\hline
\toprule
CCA (Type-1)  & ViT & \color{black}{{66.06}} & \color{black}{{67.20}} & \color{black}{{62.34}} & \color{black}{{60.47}} & \color{black}{{70.29}} & \color{black}{\textbf{75.93}} & \color{black}{{70.32}} & \color{black}{{65.76}} & \color{black}{{61.04}} \\
CCA (Type-2) & ViT & \color{black}{\textbf{69.03}} & \color{black}{\textbf{69.55}} & \color{black}{\textbf{65.92}} & \color{black}{\textbf{64.43}} & \color{black}{\textbf{72.74}} & \color{black}{{75.39}} & \color{black}{\textbf{71.89}} & \color{black}{\textbf{70.42}} & \color{black}{{63.85}} \\

\bottomrule
\end{tabular}
 }
 \vspace{-8pt}
 \caption{mAP comparisons of our methods against other studies on FashionAI. Bold: the best results among all methods. 
 Bold black: the best results among the counterparts. TF is Transformer. R50 is ResNet50. $\dagger$ indicates our reproduced results.}
\label{tab:fashionAI}
\end{table*}

%% file: exp_memory_en.tex
\subsection{Memory Efficiency} 
The ViT used in this study has 98M parameters. Individual networks are required to learn   attributes with the existing naive method, which necessitates 98M $\times K$ parameters. However, our proposed method can form multi-space embeddings with only one backbone network, thus requiring approximately 98M $ \times 1 $ parameters. As shown in \autoref{fig:fig2}, only the last layer of the ViT model is modified in the proposed CCA, and fewer than 0.1M parameters are added for conditional token embedding. Thus, the proposed method achieves SOTA performance with very few parameters, indicating high efficiency of the algorithm.

%% file: sota_en.tex
\subsection{Benchmarking}
\label{sec:eval}

\autoref{tab:fashionAI}, \autoref{tab:DARN}, and \autoref{tab:DeepFashion} present the evaluations for mAP using the metrics in \autoref{sec:Metrics}. \autoref{tab:zappos50k} shows the triplet prediction metric results. In all tables, our method outperforms the SOTA models CSN \cite{veit2017} and ASEN \cite{ma2020fine}. 
\vspace*{-10pt}
\paragraph{FashionAI} 
In \autoref{tab:fashionAI}, our method achieves SOTA performance for all categories except neck design. Overall, we achieve a +4.72\% performance improvement.

\input{sota_darn.tex}

\vspace*{-10pt}
\paragraph{DARN} 
In \autoref{tab:DARN}, the proposed model yields SOTA performance for all items. Averaged across the board, it shows a significant performance improvement of +12.15\%.

\vspace*{-10pt}
\paragraph{DeepFashion}
In \autoref{tab:DeepFashion}, the proposed model yields SOTA performance for all items. Overall, we achieve a performance improvement of +1.4\%. As shown in \autoref{tab:datasets}, although it consists of only five attributes, these contain many more classes than FashionAI and DARN at 1000, resulting in a relatively low mAP value.

\input{sota_zappos.tex}
\vspace*{-10pt}

\paragraph{Zappos50K} 
\autoref{tab:zappos50k} presents the triplet prediction metric results. Our method achieved SOTA performance, with a +3.61\% improvement compared to the previous method. Unlike the aforementioned datasets, the Zappos50K dataset is relatively simple, as indicated by the category composition in \autoref{tab:datasets} and the example in \autoref{fig:app2}.

%% file: sota_darn.tex
\begin{table*} [tb!]
\renewcommand{\arraystretch}{1.2}
\centering 
\resizebox{2\columnwidth}{!}{
\begin{tabular}{l*{11}{c}}
\toprule
\multirow{2}{*}{\textbf{Method}} &  \multirow{2}{*}{{\tb{Backbone}}} & \multirow{2}{*}{\textbf{mAP}} & \multicolumn{9}{c}{\textbf{mAP for each attribute}}  \\
\cmidrule(l){4-12}
 & & &  \tb{clothes category} & \tb{clothes button} & \tb{clothes color} & \tb{clothes length} & \tb{clothes pattern} & \tb{clothes shape} & \tb{collar shape} & \tb{sleeve length} & \tb{sleeve shape} \\
\toprule
Random baseline \cite{ma2020fine} & R50 & 32.26 & 8.49 & 24.45 & 12.54 & 29.90 & 43.26 & 39.76 & 15.22 & 63.03 & 55.54 \\
Triplet network  \cite{ma2020fine}  & R50 & 40.14 & 23.59 & 38.07 & 16.83 & 39.77 & 49.56 & 47.00 & 23.43 & 68.49 & 56.48  \\
CSN \cite{ma2020fine}               & R50 & 50.86  & 34.10 & 44.32 & 47.38 & 53.68 & 54.09 & 56.32 & 31.82 & 78.05 & 58.76\\
ASEN \cite{ma2020fine}               & R50 & {53.31} & {36.69} & {46.96} & {51.35} & {56.47} & {54.49} & {60.02} & {34.18} & {80.11} & {60.04} \\
CAMNet \cite{song2022}$^\dagger$ & R50   & 44.32 & 25.24 & 38.02 & 47.01 & 45.25 & 48.35 & 45.57 & 23.33 & 71.69 & 55.89  \\ 
M2Fashion \cite{Wan01}             & R50 & {54.29} & {36.91} & {48.03} & {51.14} & {57.51} & {56.09} & 60.77 & {35.05} & {81.13} & {62.23} \\
ASEN++ \cite{dong2021fine} & R50 & 55.94  & 40.15 & 50.42 & 53.78 & 60.38 & 57.39 & 59.88 & 37.65 & 83.91 & 60.70 \\
TF-CSN$^\dagger$  & ViT & {62.85} & {48.65} & {60.71} & {53.27} & {66.18} & {63.70} & {72.75} & {45.95} & {88.36}  & 66.35 \\
TF-ASEN$^\dagger$  & ViT & 33.52 & 6.20 & 23.28 & 31.24 & 31.37 & 41.16 & 39.02 & 15.57 & 60.88 & 54.16 \\
\toprule
\multicolumn{10}{c}{\textbf{Ours}} \\
\hline
\toprule
CCA (Type-1)  & ViT & 66.78 & 51.56 & 65.55 & 55.94 & 72.95 & 66.97 & 75.80 & 51.37 & 90.08 & \color{black}{\textbf{71.44}} \\
CCA (Type-2) & ViT & \color{black}{\textbf{68.09}} & \color{black}{\textbf{53.04}} & \color{black}{\textbf{68.21}} & \color{black}{\textbf{56.65}} & \color{black}{\textbf{74.71}} & \color{black}{\textbf{70.12}} & \color{black}{\textbf{77.03}} & \color{black}{\textbf{52.51}} & \color{black}{\textbf{90.23}} & \color{black}{{70.99}} \\

\bottomrule
 \end{tabular}
}
\centering
\vspace{-8pt}
\caption{mAP comparisons of our methods against other studies on DARN. $\dagger$ indicates our reproduced results.} 
\label{tab:DARN}
\end{table*}

\begin{table*} [tb!]
\renewcommand{\arraystretch}{1.2}
\centering 
\resizebox{2\columnwidth}{!}{
\begin{tabular}{l*{8}{c}}
\toprule
\multirow{2}{*}{\textbf{Method}} & \multirow{2}{*}{\Th{\tb{Backbone}}} & \multirow{2}{*}{\textbf{mAP}} & \multicolumn{5}{c}{\textbf{mAP for each attribute}}  \\
\cmidrule(l){4-8}
 & & & \tb{texture-related} & \tb{fabric-related} & \tb{shape-related} & \tb{part-related} & \tb{style-related} \\
\toprule
Random baseline \cite{ma2020fine} & R50 & 3.38 & 6.69 & 2.69 & 3.23 & 2.55 & 1.97 \\
Triplet network  \cite{ma2020fine}  & R50 & 7.36 &  13.26 & 6.28 & 9.49 & 4.43 & 3.33 \\
CSN \cite{ma2020fine}  & R50 & 8.01 & 14.09 & 6.39 & 11.07 & 5.13 & 3.49 \\
ASEN \cite{ma2020fine}  & R50 & 8.74 & 15.13 & 7.11 & 12.39 & 5.51 & 3.56 \\
ASEN++ \cite{dong2021fine} & R50 & 9.64 & 15.60 & 7.67 & 14.31 & 6.60 & 4.07 \\
TF-CSN$^\dagger$  & ViT & 10.04 & 15.27 & 8.11 & 14.91 & 7.40 & 4.51 \\
TF-ASEN$^\dagger$  & ViT & 8.53 & 13.98 & 6.56 & 13.39 & 5.61 & 3.13 \\
\toprule
\multicolumn{8}{c}{\textbf{Ours}} \\
\hline
\toprule
CCA (Type-1) & ViT & 10.64 & 16.18 & 8.38 & 15.98 & 7.99 & 4.78 \\
CCA (Type-2) & ViT & \color{black}{\textbf{11.04}} & \color{black}{\textbf{16.76}} & \color{black}{\textbf{8.42}} & \color{black}{\textbf{16.83}} & \color{black}{\textbf{8.47}} & \color{black}{\textbf{4.92}} \\
\bottomrule
 \end{tabular}
}
\centering
\caption{mAP comparisons of our methods against other studies on DeepFashion.} 
\label{tab:DeepFashion}
\end{table*}

%% file: sota_zappos.tex
\begin{table} [tb!]
\renewcommand{\arraystretch}{1.2}
\tiny
\centering 
\resizebox{1\columnwidth}{!}{
\begin{tabular}{lr}
\toprule
\textbf{Method} & \textbf{Prediction Accuracy(\%)}\\
\midrule
Random baseline \cite{ma2020fine}          & 50.00 \\
Triplet network \cite{veit2017} & 76.28 \\
CSN \cite{veit2017}                       & 89.27 \\
ASEN \cite{ma2020fine}                     & 90.79 \\
ADDE-C \cite{hou2021learning}                     & 91.37 \\
TF-CSN$^\dagger$                      & 94.78 \\
TF-ASEN$^\dagger$                    & 94.56 \\
\toprule
\multicolumn{2}{c}{\textbf{Ours}} \\
\hline
\toprule
CCA (Type-1)                     & {\textbf{94.98}} \\
CCA (Type-2)                    & 94.85 \\
\bottomrule
\end{tabular}
 }
 \vspace{-8pt}
 \caption{Performance of triplet prediction on Zappos50k.}
\label{tab:zappos50k}
\end{table}

%% file: ablationstudies_en.tex
\subsection{Ablation Studies}
\label{sec:ablation}

\paragraph{SOTA models applied Transformer} 
The results of the existing CSN \cite{veit2017}, ASEN \cite{ma2020fine} models are obtained with the RestNet50 as the backbone. For a fair comparison, we apply the ViT backbone rather than CNN to these methods and present the experimental results. These models are indicated as TF-CSN and TF-ASEN, respectively. To apply this to CSN and ASEN, first, CSN must accept dimensions of size $\mathbb{R}^{D}$. Hence, it must be applied in \cls $ \in \mathbb{R}^{D}$. In contrast, ASEN must accept a CNN feature map of $\in \mathbb{R}^{W \times H \times D}$ dimensions. For ViT, it must be applied in \patch $\in \mathbb{R}^{N \times D}$, which can be applied because $N$ can be reshaped to ${W \times H}$. One peculiarity is that ASEN outperforms CSN based on CNN but not that based on ViT. Overall, our proposed CCA, with the same transformer base as TF-CSN and TF-ASEN, outperforms both models.

\vspace*{-10pt}
\paragraph{Consistent Performance}
We found that previous studies yielded different performance results for the datasets. For example, \autoref{tab:fashionAI}, CAMNet \cite{song2022} outperformed ASEN and CSN, whereas, in \autoref{tab:DARN} and \autoref{tab:DeepFashion}, there are no performance results. Similarly, in \autoref{tab:DARN}, M2Fashion \cite{Wan01} outperformed ASEN and CSN, whereas, in \autoref{tab:fashionAI} and \autoref{tab:DeepFashion}, there are no results. This suggests that the performance varies with the dataset. Accordingly, we applied the CAMNet study to the DARN dataset to reproduce it. In \autoref{tab:DARN}, the $\dagger$ symbol indicates our reproduced results. CAMNet model yielded lower performance than ASEN and CSN. Moreover, ASEN outperformed CSN based on CNN in the experimental results of TF-CSN and TF-ASEN when using the transformer. This is attributed to differences in learning according to the characteristics of each dataset. Thus, learning to form embeddings for objects with multiple attributes using a single network is very difficult. In contrast, our proposed CCA consistently yields high performance for all datasets. 
\vspace*{-10pt}
\paragraph{Type-1 \vs Type-2}
These results relate to \autoref{eq:Conditionalembedding1} and \autoref{eq:Conditionalembedding2} in \autoref{sec:cca}. \autoref{tab:fashionAI} presents the results for CCA (Type-1) and CCA (Type-2); CCA (Type-2) yielded +2.97\% higher performance than CCA (Type-1). In \autoref{tab:DARN} and \autoref{tab:DeepFashion}, CCA (Type-2) showed +1.31\% and +0.4\% higher performance, respectively. In \autoref{tab:zappos50k}, CCA (Type-1) yielded +0.42\% higher performance. CCA (Type-2) was slightly higher in the previous three benchmark sets, whereas CCA (Type-1) was slightly higher by 0.13\% in this dataset. However, CCA (Type-1) and CCAN (Type-2) outperformed all results of the previous studies and TF-CSN and TF-ASEN described above, achieving SOTA performance. 

%% file: conclusion_en.tex
\section{Conclusion}
This study investigates forming embeddings for an object with multiple attributes using a single network, which is generally difficult in practice. However, the proposed method can extract various specific attribute features using a single backbone network. The proposed network enables multi-space embedding for multiple attributes. Finally, our proposed algorithm achieved SOTA performance in all evaluation metrics for the benchmark datasets.

%% file: supp_en.tex
\clearpage

\begin{center}
\textbf{\large Supplementary material for \\ ``Conditional Cross Attention Network for Multi-space Embedding without Entanglement in Only a SINGLE Network''}
\end{center}

\maketitle
\thispagestyle{empty}



\appendix

\renewcommand{\theequation}{A\arabic{equation}}
\renewcommand{\thetable}{A\arabic{table}}
\renewcommand{\thefigure}{A\arabic{figure}}


\section{Datasets}
\label{sec:data_examples}

\paragraph{FashionAI \cite{Zou2019FashionAIAH}}
The data published in the FashionAI Global Challenge 2018 has 180,335 apparel images. This dataset comprises 8 fashion attributes containing 55 classes each. 
\vspace*{-15pt}
\paragraph{DARN \cite{Huang01}} 
An open dataset for attribute classification and street-to-shop image retrieval, comprising 253,983 images and 9 attributes. Each attribute contains 185 classes. The data is provided as image URLs; excluding broken URLs that cannot be downloaded, we used 195,771 URLs. 
\vspace*{-15pt}
\paragraph{DeepFashion \cite{liu2016deepfashion}}
This dataset comprises 289,222 images and 6 attributes. Each attribute contains 1000 classes. 
\vspace*{-26pt}
\paragraph{Zappos50K \cite{Zappos}}
This dataset comprises 50,025 shoe images collected from Zappos.com. It consists of 4 attributes containing 34 classes each.

\begin{figure}
\begin{center}
\includegraphics[width=1.0\linewidth]{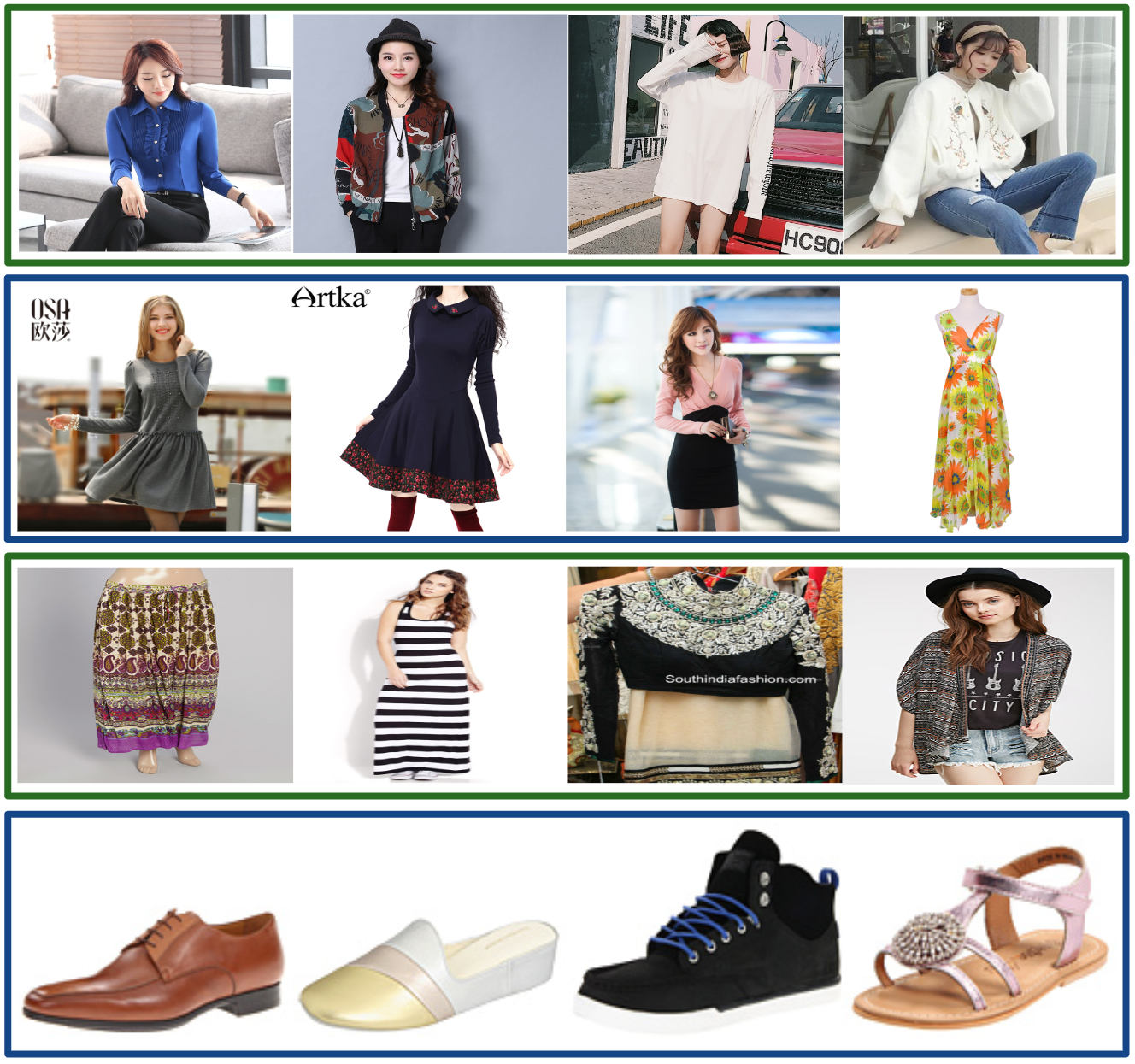}
\end{center}
   \vspace*{-15pt}
   \caption{ Examples of Our TrainSets. The order of each row is FashionAI, DARN, DeepFasion and Zappos50K.}
\label{fig:app2}
\end{figure}

\autoref{fig:app2} presents actual examples using the four training sets. The figure shows four examples in the order of FashionAI \cite{Zou2019FashionAIAH}, DARN \cite{Huang01}, DeepFashion \cite{liu2016deepfashion}, Zappos50k \cite{Zappos}. 




\section{More experiments}
\label{sec:more_exp}

\subsection{Benchmarking : DeepFashion}
\label{sec:more_DeepFashion}

\autoref{tab:DeepFashion} presents the experimental results for DeepFashion, described in detail in \autoref{sec:eval}. 


\section{More Visualization}
\label{sec:more_vis}

\subsection{Ranking and Attention Heat map}
\label{sec:more_ranking}

\begin{figure*}
\begin{center}
\includegraphics[width=1.0\linewidth]{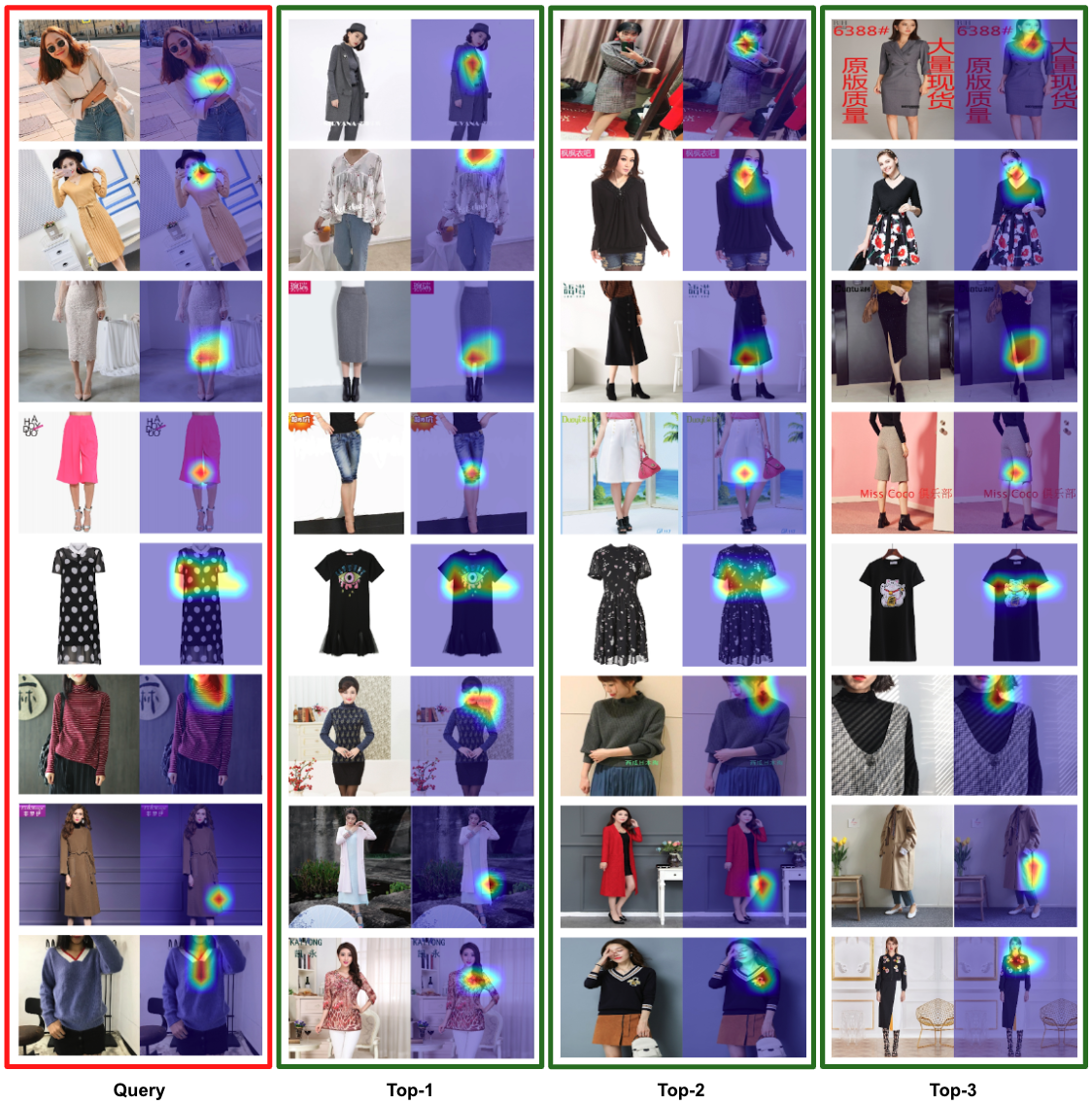}
\end{center}
   \vspace*{-15pt}
   \caption{ Examples of our top 3 ranking  pair (image, attention heat map) results for FashionAI of 8 attributes. Red rectangle is query images. The order of each line is lapel design (notched), neckline design (round), skirt length (floor), pant length  (midi), sleeve length (short), neck design (low turtle), coat length (midi) and collar design (peter pan).}
\label{fig:app3}
\end{figure*}

\autoref{fig:app3} shows the Top 3 results along with each actual attention map. Each part of each attribute is considered, interpreted as the result of disentanglement multi-space modeling. The order in the figure is lapel design (notched), neckline design (round), skirt length (floor), pant length (midi), sleeve length (short), neck design (low turtle), coat length (midi), and collar design (peter pan).

\subsection{Ours \vs Previous Works : Multi-Space Embedding}
\label{sec:more_previouswork}

\begin{figure*}
\begin{center}
\includegraphics[width=1.0\linewidth]{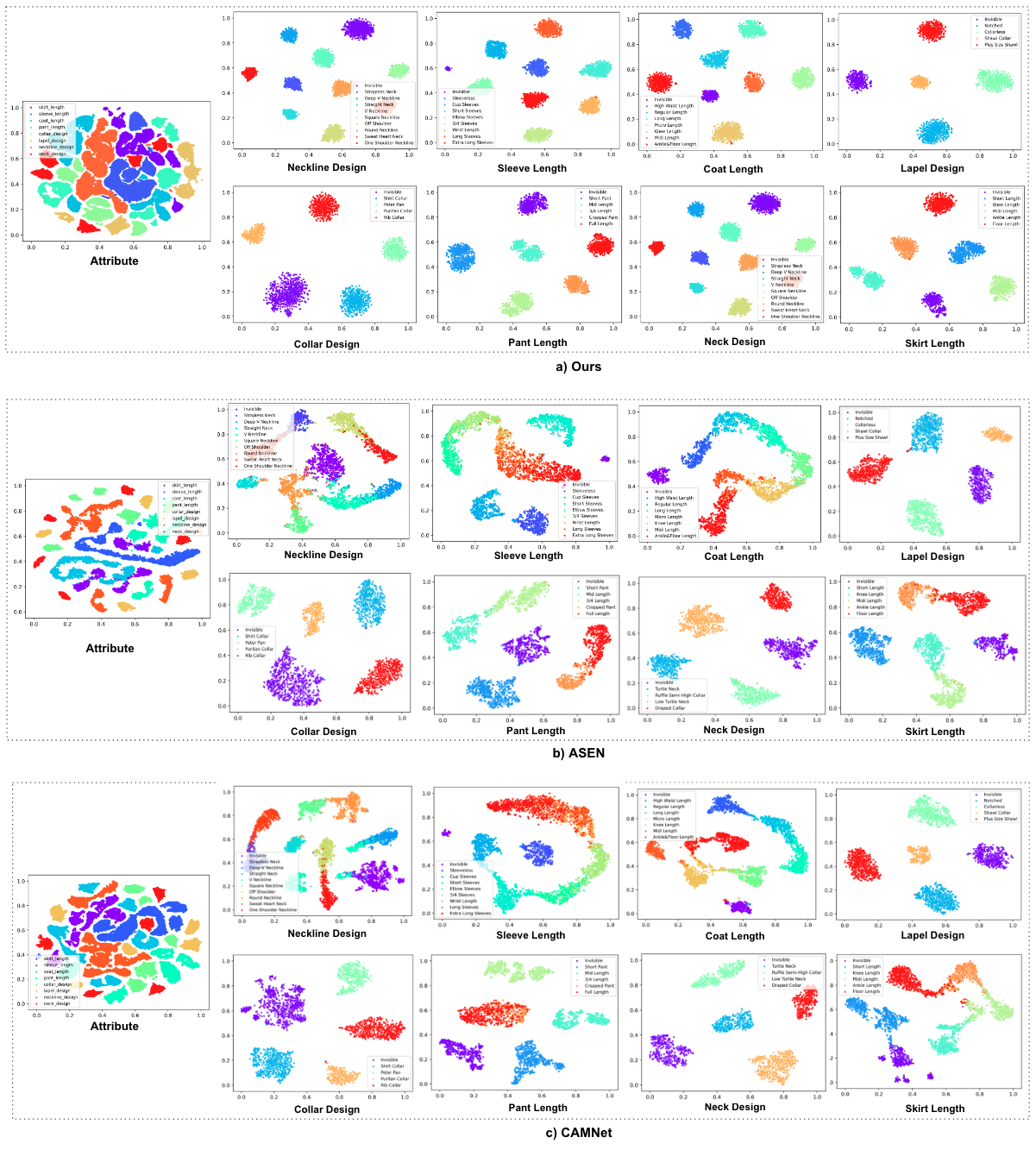}
\end{center}
   \vspace*{-15pt}
   \caption{ Ours \vs Previous Works (ASEN, CAMNet) : Multi-space embedding's visualization using t-SNE about FashionAI.}
\label{fig:app4}
\end{figure*}

\autoref{fig:fig7} comparatively analyzed the embedding results of our study and previous studies. Of the 8 categories, the results for Neck Design, Sleeve Length, and Coat Length were presented. \autoref{fig:app4} shows the expanded results for all 8 categories. Our method solves the entanglement problem much better than ASEN \cite{ma2020fine} and CAMNet \cite{song2022}.


\algrenewcommand{\algorithmiccomment}[1]{$\triangleright$ #1}
\begin{algorithm}[t!]
\small
\caption{Pseudo-Code for CCA Training}
\label{alg_train}
\begin{algorithmic}[1]
\State \textbf{input:} 
Image $\mathcal{I}$, Condition $c$ \\
batch  $\mathcal{B}$, training epochs $K$, triplet set $\mathcal{T}$ \\
Self Attention Block $SA$, Conditional Cross Attention $CCA$
\For{$epoch= 1,...,K $}
    \For{$\mathcal{B}=1,...,M \in \mathcal{T} $}
        \State $Triplet(\mathcal{A}_c, \mathcal{P}_c, \mathcal{N}_{c})\gets \mathcal{B}$
        \State $\mathcal{I}, c \gets \mathcal{A}_c, \mathcal{P}_c, \mathcal{N}_{c}$
        \State $\mathcal{Q}_i, \mathcal{K}_i, \mathcal{V}_i \gets Token\_{Embedding}(\mathcal{I})$
        \For{$l=1,...,(\mathcal{L}-1) $}  
            \State $ \mathcal{Q}_i, \mathcal{K}_i, \mathcal{V}_i \gets SA (\mathcal{Q}_i, \mathcal{K}_i, \mathcal{V}_i)$               
        \EndFor    
        \State \textbf{Last iteration} $ l = \mathcal{L} $ \textbf{do} 
        \State \hspace{\algorithmicindent} $ \mathcal{Q}_c \gets Conditional\_Token\_Embedding(c) $
        \State \hspace{\algorithmicindent} $ \cls \gets CCA (\mathcal{Q}_c, \mathcal{K}_i, \mathcal{V}_i)$
        \State \hspace{\algorithmicindent} $ f \gets l2(FC(\cls))$
        \State calculate $ f_a, f_p, f_n \gets Triplet(\mathcal{A}_c, \mathcal{P}_c, \mathcal{N}_{c})$
        \State calculate triplet loss $\mathcal{L} (f_a, f_p, f_n | c)$
        \State calculate gradients of $\nabla\mathcal{L}(\theta)$
        \State $\theta \gets Adam(\nabla\mathcal{L}(\theta)$)
    \EndFor
\EndFor
\end{algorithmic}
\end{algorithm}